\documentclass{article}

\usepackage{nips15submit_e,times}
\usepackage{subcaption}
\usepackage{verbatim}
\usepackage{graphicx}
\usepackage{amsmath, amssymb, amsthm}
\usepackage{url}
\usepackage[numbers, sort]{natbib}

\definecolor{darkred}{rgb}{0.5,0,0}
\definecolor{darkgreen}{rgb}{0, 0.3,0}
\definecolor{darkblue}{rgb}{0,0,0.6}
\definecolor{LightGray}{rgb}{.6,.6,.6}
\definecolor{LightLightGray}{rgb}{.9,.9,.9}

\usepackage{bbm}

\newcommand{\defn}[1]{\textbf{#1}}
\newcommand{\defas}{:=}

\newcommand{\Nats}{\Naturals}

\newcommand{\iid}{\stackrel{\text{iid}}{\sim}}
\newcommand{\ind}{\stackrel{\text{ind}}{\sim}}
\newcommand{\simas}{\stackrel{\text{a.s.}}{\sim}}

\def\EM#1{\ensuremath{#1}}
\def\mbb#1{\EM{\mathbb{#1}}}

\def\Naturals{{\EM{{\mbb{N}}}}}

\newcommand{\mbo}{\mathbbm{1}}

\theoremstyle{plain}

\theoremstyle{definition}

\theoremstyle{remark}

\title{Completely random measures for modeling\\ power laws in sparse graphs}

\author{Diana Cai \\ Department of Statistics \\ University of Chicago \\
    \texttt{dcai@uchicago.edu}
\And Tamara Broderick \\ Department of EECS \\ Massachusetts Institute of Technology \\
    \texttt{tbroderick@csail.mit.edu}}

\begin{document}

\nipsfinalcopy
\maketitle

\begin{abstract}
Network data appear in a number of applications, such as online social networks
and biological networks, and there is growing interest in both developing models
for networks as well as studying the properties of such data. Since individual network datasets continue to grow in size, it is necessary to develop models that accurately represent the real-life scaling properties of networks. One behavior of interest is having a power law in the degree distribution. However, other types of power laws that have been observed empirically and considered for applications such as clustering and feature allocation models have not been studied as frequently in models for graph data. In this paper, we enumerate desirable asymptotic behavior that may be of interest for modeling graph data, including sparsity and several types of power laws. We outline a general framework for graph generative models using completely random measures; by contrast to the pioneering work of Caron and Fox (2015), we consider instantiating more of the existing atoms of the random measure as the dataset size increases rather than adding new atoms to the measure. We see that these two models can be complementary; they respectively yield interpretations as (1) time passing among existing members of a network and (2) new individuals joining a network. We detail a particular instance of this framework and show simulated results that suggest this model exhibits some desirable asymptotic power-law behavior.
\end{abstract}

\section{Introduction}

In recent years, network data has increased in both ubiquity and size.
As network data appear in a growing number of applications---such
as online social networks, biological networks, and networks representing
communication patterns---there
is growing interest
in developing models and inference for such data and studying its properties.
Bayesian generative models for network data include, but are not limited to,
the stochastic block model \citep{MR718088} and
variants \citep{C.Kemp:2006:53fd9, DBLP:conf/mlg/XuTYYK07},
the infinite relational model \citep{C.Kemp:2006:53fd9},
the latent feature relational model \citep{DBLP:conf/nips/MillerGJ09},
the infinite latent attribute model \citep{DBLP:conf/icml/PallaKG12},
and the random function model \citep{DBLP:conf/nips/LloydOGR12}.

Crucially, individual network data sets also continue to
increase in size. Thus, it is not enough to develop models for networks, but
in particular it is necessary to develop
models that accurately represent the real-life scaling properties of
networks. As particular networks increase in size, it has already become
apparent that all of the models listed above share at least one undesirable
scaling property. In particular, they all fit the assumptions of the
\emph{Aldous--Hoover
Theorem} \citep{MR637937,Hoover79},
which implies that they all generate \emph{dense} graphs with
probability one \cite{DBLP:journals/pami/OrbR14}.
Here, we say a graph is dense if the number of edges
in the graph grows asymptotically as the square of the number of vertices in the graph.
By contrast, we say a graph is \emph{sparse} if the number of edges grows
sub-quadratically as a function of the number of vertices in the graph.

This disconnect between desired asymptotic behavior and
model specifications motivates the development of new models that
achieve sparsity rather than always generating dense graphs.
Some initial work in this direction has been pioneered by
\citet{caron2014arxiv}. But just as a wide variety of models with
a wide variety of behaviors are available for other applications,
it is desirable to populate a broader toolbox of appropriate models
for network data.

It remains to cultivate a more complete understanding of what behaviors are
desirable in network data. Dense graphs are generally considered
undesirable. But graph dense-ness describes just one potential graph
behavior. Scaling behavior has been much more extensively studied
in other domains.
Since power laws are widely observed in real data
\citep{DBLP:journals/im/Mitzenmacher03, newman2005power,
DBLP:journals/siamrev/ClausetSN09},
these investigations into scaling have tended to focus on power law
behaviors.
For instance, \citet{MR2318403} have
thoroughly characterized a wide variety of power laws that may be exhibited
in clustering models and have, moreover, shown that many of these power laws
are equivalent. One behavior of interest is a power law
in the number of clusters as the
number of data points grows. This
behavior is roughly
analogous to considering a power law in the number of
edges as the number of vertices grows. But \citet{MR2318403}
consider a much wider range of potential power laws. Likewise,
\citet{MR2934958} have enumerated
a range of power laws for \emph{feature allocations},
a generalization of clustering where each data point may belong
to any non-negative integer number of groups---now called
features instead of clusters. While
some authors have drawn connections between
feature allocations and certain types of networks \cite{DBLP:conf/nips/Caron12}, an
exhaustive enumeration of asymptotic network
behaviors of interest in graph data---beyond the simple
divide between dense and sparse graphs---is still missing.

Not only have previous authors studied power laws
for clustering and feature allocations, but they have
detailed particular, practical generative models for
achieving these power laws---and these models
typically lead to corresponding inference algorithms as well.
For instance, the canonical power law model
for clustering is the Pitman--Yor process
\cite{pitman1997two,goldwater2005interpolating,teh2006hierarchical},
and the canonical power-law model for feature allocations
is the three-parameter beta process \cite{teh2009indian,MR2934958}.

Below we outline a general framework for graph generative models
in Section~\ref{sec-model}.
We detail a particular instance
of our framework that gives a (new) generative model for networks
that may be applied
in practice.
We develop a list of asymptotic behaviors of
interest in network models in Section~\ref{sec-powerlaw}.
In Section~\ref{sec-simulations}, we show preliminary results that suggest this model
exhibits some desirable asymptotic (power-law) behavior. And we
suggest empirical, theoretical, and algorithmic developments for future research
in Section~\ref{sec-future}.

\section{Generative framework}
\label{sec-model}

We first consider a general framework for generating graphs.
Then we consider a specific case using completely random measures.
Lastly, we show how this can be used to create models in practice by consider a
beta process as the underlying completely random measure.

\subsection{Motivation}

Let $(\Omega, \Sigma, \mathbf{P})$ be a probability space.
A \defn{random measure} $W$ on $(\Omega, \Sigma)$ is a random measure-valued
element such that
$W(A)$ is a random variable for any measurable set $A \in \Sigma$.

Now suppose $W$ is an atomic random measure with
atoms $(\theta_i)_{i=1,\ldots,M}$ and weights $(w_i)_{i=1,\ldots,M}$,
where for all $i$, we have $w_i \in (0,1)$, that is,
\begin{align*}
W = \sum_{i=1}^{M} w_i \delta_{\theta_i},
\end{align*}
where $M$ may be random or infinite.
In this case, we can use the weights to generate the adjacency matrix $G$
of a graph by
independently drawing edges ($G_{ij}=1$) with probability $w_i w_j$.
Given $W$, we can draw a multigraph, in which edges can have multiplicity, by
drawing edges independently and identically $N$ times.

We imagine each $\theta_i$ as corresponding to a vertex. So $w_i w_j$
is the probability of an edge forming between the vertices corresponding
to $\theta_i$ and $\theta_j$.
Thus, if $M$ is infinite, we theoretically have a countably infinite vertex set.
However, another perspective is that only vertices that participate in some
edge count toward the total number of vertices;
we call the vertices that are connected via any edge \defn{effective vertices}.
From this perspective, having
an infinite latent vertex collection ($M = \infty$) is necessary to allow the number of
effective vertices to grow without bound.
Completely random measures provide an option for generating a countably infinite number of
atoms in our random measure $W$.

A \defn{completely random measure}
$W$ on $(\Omega, \Sigma)$ is a random measure with the additional requirement
such that for any finite, disjoint measurable sets
$A_1,\ldots,A_n \in \Sigma$, the random variables $W(A_1),\ldots,W(A_n)$
are (pairwise) independent \cite{MR1207584}.
Completely random measures can be constructed from a Poisson point process with rate
measure $\nu(d\theta, dw)$ in the following way: if we draw a sample
$(\theta_i, w_i)_{i \in \Nats}$ from a Poisson point process,
we construct $W$ as follows:
\begin{align*}
	W = \sum_{i=1}^{\infty} w_{i} \delta_{\theta_{i}}.
\end{align*}
All completely random measures can be obtained in this way (along with a
deterministic component and a fixed atomic component) \cite{MR1207584}.

\subsection{Generative model}

Let $W$ be a draw from the Poisson point process component of a completely random
measure, where we assume that $\nu(d\theta, dw)$ has support on $w \in (0,1)$.
To sample a graph given $W$, we
draw
an edge connection $C_{n,i,j} \ind \text{Bernoulli}(w_i w_j)$,
for $n = 1,\ldots,N$,
resulting in $G_{N,i,j} = \sum_{n=1}^{N} C_{n,i,j}$ edges for each pair of vertices $(i,j)$.
For simplicity, we assume there are no loops; however, it is straightforward to
adapt the model to include loops.
In this paper, we primarily consider the restriction of $G_N$ to a binary array $Z_N$,
where $Z_{N,i,j} \defas \min\{G_{N, i,j},1\}$.

This generative model can have the following interpretation:
$N$ can be seen
as ``time passing," where as $N$ grows, more links are being generated in the
network, thus bringing in more edges as well as effective vertices (i.e., vertices
that are connected to at least one other vertex).

\subsection{Example using beta processes}

The beta process is an example of a completely random measure with rate measure
\begin{align*}
    \nu(d\theta,dw) = cw^{-1} (1-w)^{c-1} dw \, B_0(d\theta),
\end{align*}
where $c>0$ is the concentration parameter and $B_0$ is the base measure.
It is known that for feature allocation applications, the beta
process does not give power law behavior in scaling of quantities such as the
number of instantiated features.

However, an extension of the beta process, the three-parameter beta process, is
known to give power laws in feature allocation
\citep{teh2009indian,MR2934958}.
The three-parameter beta process
has rate measure
$\nu$, a $\sigma$-finite measure with density
\begin{align}
    \nu(d\theta,dw) = \frac{\Gamma(1+c)}{\Gamma(1-\alpha)\Gamma(c+\alpha)}
    w^{-1-\alpha} (1-w)^{c +\alpha-1} dw\, B_0(d\theta),
\end{align}
where $\alpha \in (0,1)$ and $c > -\alpha$.
We denote a draw from the three-parameter beta process as
$W\sim\text{BP}(\theta,\alpha,B_0)$.

To obtain power law behavior in graphs, we are similarly interested in completely
random measures which can produce such behavior.

\begin{figure}[t]
\centering
    \centering
    \begin{subfigure}{0.48\linewidth}
        \centering
        \includegraphics[scale=0.35]{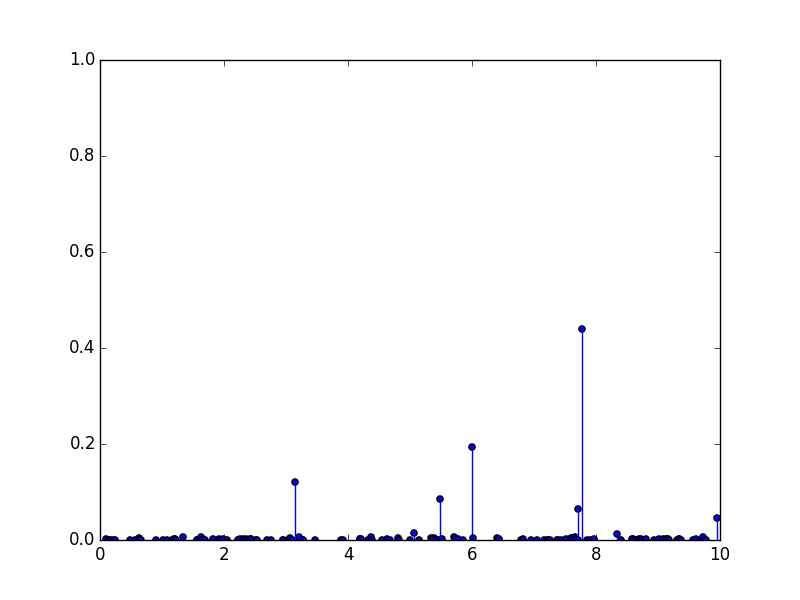}
        \caption{Realization $W \sim \text{PP}(\nu)$}
    \end{subfigure}
    \begin{subfigure}{0.42\linewidth}
        \centering
        \includegraphics[scale=0.45]{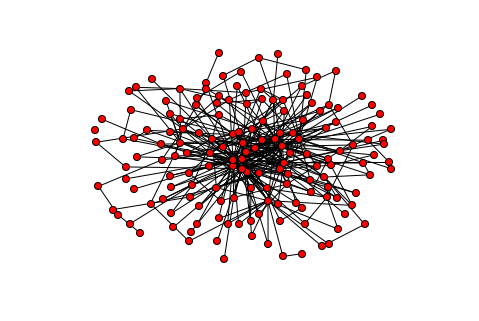}
        \caption{A sample}
    \end{subfigure}
    \caption{A realization of $W$ when the underlying completely random measure is a three-parameter beta
        process, and the resulting binary graph $Z_N$ sampled given $W$.
        Here $N=100$ and
        the parameters of the beta process are $(\theta, \alpha, \gamma)$ = (1,
                0.5, 5), where $\theta$ is the concentration parameter
            $\alpha$ is the discount parameter, and $\gamma$ is the Poisson
            rate parameter.
        }
\end{figure}

\section{Power laws for graphs}
\label{sec-powerlaw}

It remains to be seen whether
graph models produced under this framework
exhibit desirable asymptotic and power law behaviors.
For instance,
we may be interested in graph behaviors similar to known power laws in partitions
\cite{MR2318403}
and feature allocations \cite{MR2934958}.
In what follows, we define a number of power laws that may be of interest
in graph modeling. In future work, we aim to characterize under what
circumstances models fitting the framework from
the previous section exhibit these power laws.

\paragraph{Graph quantities.} We first establish a few quantities of interest
for which we want to define power laws.
Note that all edge counts are considered with respect to the undirected
graph given by the binary array $Z_N$, and
for simplicity, we assume there are no loops;
it is straightforward to adapt this section to graphs with loops.

We define the \defn{degree} $D_{N,i}$ of a vertex $i$ to be
\begin{align*}
    D_{N,i} \defas \sum_{i \in \Nats} Z_{N,i,j} = \sum_{i \in \Nats}
    \mbo\{G_{N,i,j} > 0\},
\end{align*}
i.e., the number of edges vertex $i$ is connected to.

A vertex $i$ has a triangle if, when $i$ is connected to a vertex $j$ and a vertex
$k$, there is also an edge between $j$ and $k$.
Let $T_{N,i}$ denote the \textbf{number of triangles} that $i$ participates in, i.e.,
\begin{align*}
    T_{N,i} := \sum_{j,k \in \Nats} \mbo\{ Z_{N,i,j} = 1\}  \mbo\{ Z_{N,i,k} =
    1\} \mbo\{ Z_{N,j,k} =1\}.
\end{align*}

We define the \defn{effective number of vertices} $\mathcal{V}_N$ to be the total number of
vertices with nonzero degree:
\begin{align*}
    \mathcal{V}_N \defas \sum_{i\in\Nats} \mbo\{D_{N,i} > 0\}.
\end{align*}
Note that in our definition, the number of vertices reflects the effective number
of vertices in the network, rather than the size of the (infinite) adjacency matrix.

The \textbf{total number of edges} $\mathcal{E}_N$ is given by
\begin{align*}
    \mathcal{E}_N := \frac{1}{2} \sum_{i \in \Nats} D_{N,i}
\end{align*}

We define the quantity
\begin{align*}
    \mathcal{D}_{N,r} \defas \sum_{i \in \Nats} \mbo\{D_{N,i} = r\},
\end{align*}
which gives the number of vertices with exactly degree $r$,
and
\begin{align*}
    \mathcal{T}_{N,r} \defas \sum_{i\in\Nats} \mbo\{T_{N,i} = r\},
\end{align*}
which gives the number of vertices with exactly $r$ triangles.

We now characterize several types of power laws that may occur in graphs.

\textbf{Type I power law}:
We first consider a power law in the asymptotic number of edges as a function of the
asymptotic number of (effective) vertices.

\begin{align*}
\mathcal{E}_N \simas c\,\mathcal{V}_N^a, \, N \rightarrow \infty,
\end{align*}
for constants $c > 0, a \in (1,2)$.

We might also consider power laws in the counts of vertices and triangles.

\textbf{Type IIa power law}: For instance, we might have a power law in the number of vertices of a
certain degree.
\begin{align*}
\mathcal{D}_{N, r} \simas c\, \mathcal{V}_N^a, \,     N \rightarrow \infty
\end{align*}

\textbf{Type IIb power law}: Similarly, we might have a power law in the number
of vertices with a certain number of triangles
\begin{align*}
    \mathcal{T}_{N, r} \simas c \, \mathcal{V}_N^a, \, N \rightarrow \infty
\end{align*}
Note that $c$ and $a$ need not be the same constants across these power laws.

These two types of power laws have behavior similar to Heaps' law
\cite{heaps1978information}
and Zipf's law \cite{zipf1949human} and have been studied extensively in a variety of
real-world data.
Some examples of graphs with this type of power law include the number of hyperlinks
in relation to the number of users (and other variables)
in an internet graph \cite{DBLP:journals/im/Mitzenmacher03} and web caching strategies for
the number of requests for webpages \cite{adamic2002zipf}.

A fundamentally different type of power law reminiscent of the kind defined by
\citet{MR2934958}
for feature modeling is given by the distribution of a quantity on a vertex,
rather than the asymptotic values of counts; the next type of power law gives a
power law in the degree and triangle distribution.

\textbf{Type IIIa power law}:
A power law for the degree distribution at a vertex $i$ is given by:
\begin{align*}
    \Pr(D_{N,i} > M) \sim c M^{-a}.
\end{align*}
\textbf{Type IIIb power law}: Similarly, we could consider the number of triangles at a vertex $i$:
\begin{align*}
    \Pr(T_{N,i} > M) \sim c M^{-a}.
\end{align*}
These types of power laws have been widely studied in
a number of real-world graphs, such as
degrees of proteins in a protein-interaction network of yeast, degrees
of metabolites in the metabolic network of E.\ coli; see
\citet{DBLP:journals/im/Mitzenmacher03, newman2005power,
DBLP:journals/siamrev/ClausetSN09} for more details.
Triangle distribution power laws have been observed in a number of real world
social networks, such as LinkedIn and Twitter, and YahooWeb
\citep{DBLP:conf/pakdd/KangMF11}.

\section{Simulations}
\label{sec-simulations}

In this section, we explore the behavior of graphs generated by this model
via simulation.
In particular, we are interested in seeing if the model can produce sparse
graphs and whether it exhibits any of the power laws described in
Section~\ref{sec-powerlaw}.
We consider the case when the underlying completely random measure is the
three-parameter beta process,
i.e., we draw a realization $W\sim\text{BP}(\theta,\alpha,B_0)$ according to the
stick breaking representation given in \citet*{MR2934958}:
\begin{align*}
W             & = \sum_{i=1}^\infty \sum_{j=1}^{C_i} V_{i,j}^{(i)}
    \prod_{l=1}^{i-1} (1 - V_{i,j}^{(\ell)})\delta_{\psi_{i,j}} \\
C_i           & \iid \text{Pois}(\gamma) \\
V_{i,j}^{(\ell)} & \ind \text{Beta}(1-\alpha, \theta + \ell\alpha) \\
    \psi_{i,j} &\iid \frac{1}{\gamma} B_0.
\end{align*}

For the beta process we truncated the number of rounds to 5000, i.e., we drew
5000 Poisson random variables $C_i$.
The parameters of the beta process were set as follows:
$\gamma=3, \theta=1, \alpha=0.1$.
The number of Bernoulli draws $N$ was varied at $N=50,\ldots,2000$ at
increments of $10$.

\begin{figure}[t]
\centering
\begin{subfigure}{0.49\linewidth}
    \centering
    \includegraphics[scale=0.5]{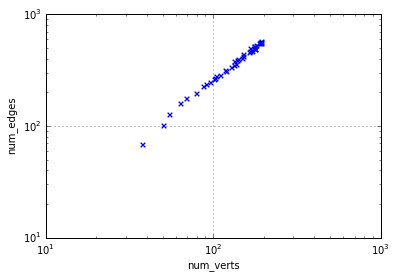}
    \caption{Scaling of vertices $V_N$ vs edges $E_N$}
    \label{fig-type-i}
\end{subfigure}
\begin{subfigure}{0.49\linewidth}
    \centering
    \includegraphics[scale=0.5]{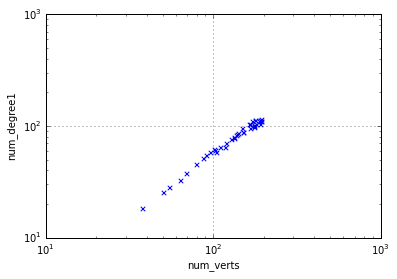}
    \caption{Scaling of vertices with degree 1 $D_{N,1}$}
    \label{fig-type-ii}
\end{subfigure}
\begin{subfigure}{\linewidth}
    \centering
    \includegraphics[scale=0.5]{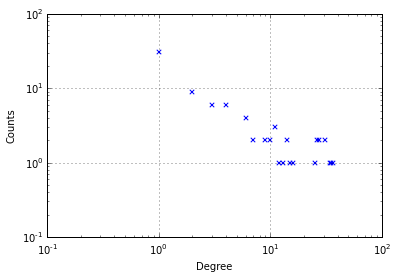}
    \caption{Degree distribution, $N=100$}
    \label{fig-degree}
\end{subfigure}
    \caption{Simulated results of scalings of vertices and edges, the scaling of
    the vertices with number of degree 1 vertices, and the degree distribution
    of a single graph with $N=100$.}
    \label{beta-scaling}
\end{figure}

In Figure~\ref{beta-scaling},
we show preliminary results from our simulations.
Figure~\ref{fig-type-i} shows the scaling of the number of vertices $V_N$
with the number of edges $E_N$.
From this plot,
we see that the model produces sparse behavior in graphs, as it is sub-quadratic
in the scaling between the number of vertices and the number of edges.
We examined the potential of having a type I power law
by fitting a line to the higher number of vertices, which gave a slope of $1.2$.
Thus, from our simulations, it appears that
the scaling between the number of vertices and the number of edges
follows a Type I power law.

In Figure~\ref{fig-type-ii}, we show the scaling between the number of vertices
$V_N$ and the number of vertices with degree 1 $D_{N,1}$.
We checked the slope of the larger vertices, and found
that these points had a slope of $1.1$.
Thus, this simulation shows the appearance of
a Type IIa power law relationship between the two quantities.

Lastly, we plot the degree distribution for a single graph when $N=100$ in
Figure~\ref{fig-degree}.
For this plot, we checked the slope of the lower degrees and found
a slope of $-1.6$; thus, it appears
that this model produces
a type IIIa power law in the degree distribution.

From our preliminary results, it seems promising that our framework generates
graphs exhibiting several types of power laws.
In future work, we will examine the behavior of other types of power laws, e.g.,
Type IIb and Type IIIb, and it remains to prove the asymptotic properties of
our model.

\section{Future directions}
\label{sec-future}

We have described a framework for generating graphs using completely random
measures and have characterized various types of power laws that may be
desirable in a network model.
In future work, we will study additional empirical power laws that can be
obtained using a three-parameter beta process as well as other completely
random measures.
An important next step is to study the asymptotic scalings of quantities
and
distributions in graphs produced from this model. Here, we are interested in
proving whether this model can produce certain power laws or whether it can be
shown that the model does not produce those power laws.
In addition to the types of power laws we examined empirically, in our theoretical
analysis, we will investigate whether this model can also capture other kinds of
power laws, such as the ones described in Section~\ref{sec-powerlaw}.
Another direction is to fit the model using an efficient inference algorithm
for this model on real-world networks.
There are additional modeling extensions to explore, such as modeling block
structure and sequential modeling (e.g., for triangles).

\bibliographystyle{plainnat}
\bibliography{sources}

\end{document}